\pdfoutput=1

\documentclass[11pt]{article}

\usepackage[preprint]{acl}
\usepackage{listings}

\usepackage{times}
\usepackage{latexsym}
\usepackage{tikz}
\usepackage{subcaption}
\usepackage{amsmath}   
\usepackage{amssymb}   
\usetikzlibrary{positioning, arrows.meta, shapes.multipart, fit}
\usepackage{float}
\usepackage[dvipsnames]{xcolor}

\usepackage[T1]{fontenc}

\usepackage[utf8]{inputenc}

\usepackage{microtype}

\usepackage{inconsolata}

\usepackage{graphicx}
\usepackage{amsmath}
\usepackage{enumerate}
\usepackage{array}
\usepackage{booktabs}
\usepackage{enumitem}
\usepackage{fontawesome}
\usepackage{xcolor}
\usepackage{multirow}
\definecolor{darkspringgreen}{rgb}{0.09, 0.45, 0.27}
\usepackage{orcidlink}

\title{What the ``Spotless'' Mind Remembers: How Knowledge Entanglement Shapes What Leaks After Unlearning in LLMs}

\author{
Aakriti Shah \\
Computer Science\\
University of Southern California \\
\texttt{shahaakr@usc.edu}
\And
Yifan Hu \\
Computer Science\\
Northeastern University \\
\texttt{yif.hu@northeastern.edu}
\And
Thai Le \\
Computer Science\\
Indiana University \\
\texttt{tle@iu.edu}
}

\begin{document}
\maketitle

\begin{abstract}
Unlearning in large language models (LLMs) is usually evaluated as whether an ``unlearned'' fact can be recovered. We instead ask whether a fact's structural entanglement with the rest of a model's knowledge predicts whether it leaks after unlearning, whether this relationship changes systematically, and whether it is causal. Across varied unlearning algorithms (WHP and GA+KL), two domains (both fictional \textit{Harry Potter} and non-fictional \textit{U.S. Senators, 2000-2010}), and four models (2.7B-13B parameters), we find that before unlearning, more entangled facts are recalled more often (r = +0.39 to +0.51). WHP weakens this relationship but stays positive (r = +0.16 to +0.33); GA+KL inverts it in every domain and model size (r = -0.14 to -0.25). To our knowledge, this is the first report of this specific reversal in the unlearning literature. To confirm this is causal beyond correlation, we directly manipulate a prompt's entanglement score, holding its content, target model fixed, and show recall moves in the predicted direction and reverses sign under GA+KL in the same direction as the correlational analysis. This manipulation is our central evidence that unlearning acts on the underlying knowledge structure, not just its output. Building on this, we train a predictive model that estimates a prompt's post-unlearning factuality/hallucination profile before unlearning is run, creating a unique way of model auditing in triaging which prompts are likely to leak.
\end{abstract}

\section{Introduction}

\begin{figure}[tb!]
    \hspace{-0.7cm}
    \includegraphics[width=8.3cm]{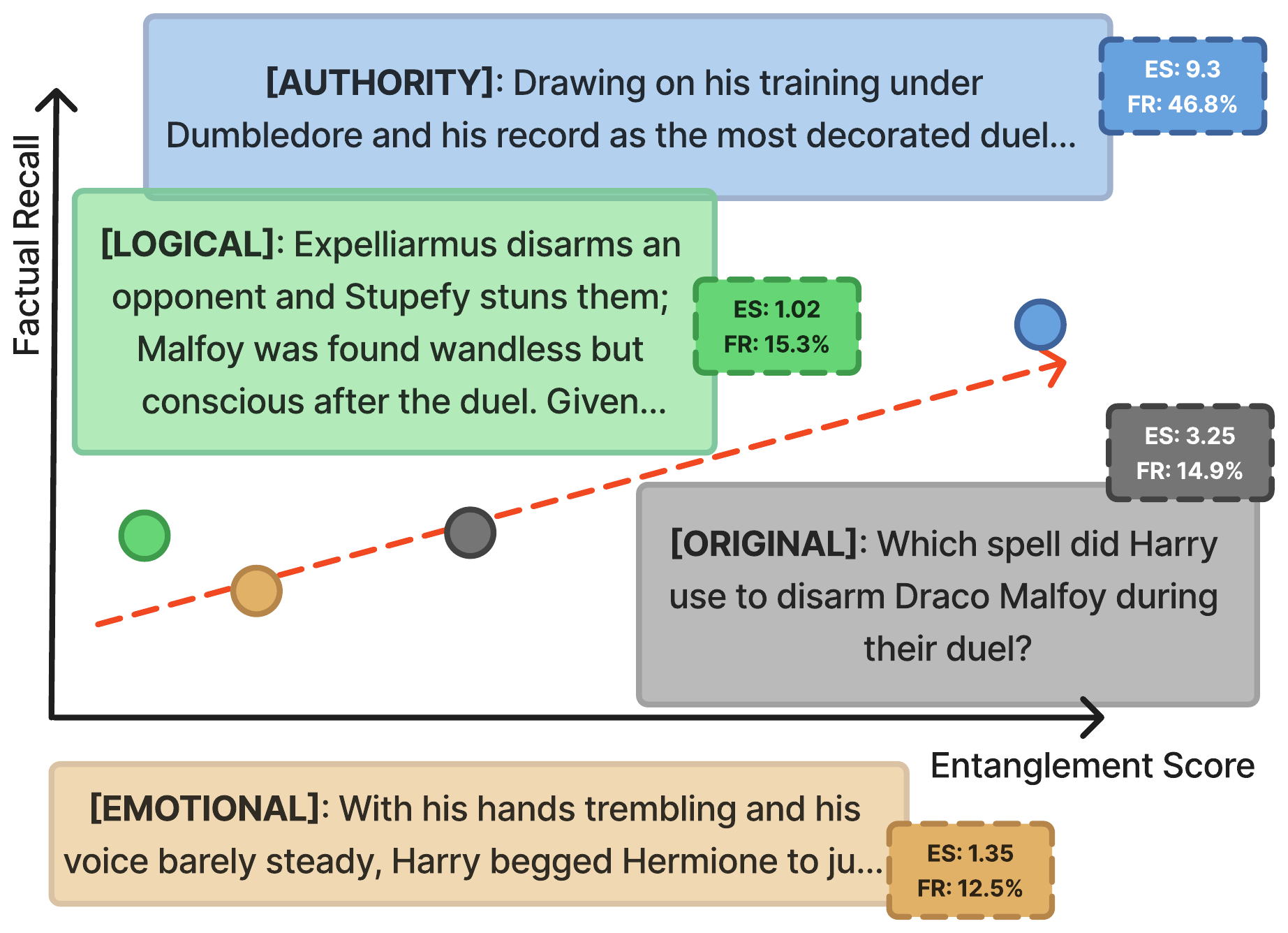}
    \vspace{-0.7cm}
    \caption{{\sc SKeB} models the relationship between (1) \textbf{knowledge entanglement} in prompt content, (2) \textbf{how the prompt is delivered}, and (3) \textbf{unlearned LLMs' behavior}. We find consistent correlations among all three, with sign depending on the unlearning algorithm.}
    \label{fig:image}
    \vspace{-0.6cm}
\end{figure}

If machine \textit{learning} draws its name from human cognition, can machine \textit{unlearning} reflect how knowledge is forgotten? Unlearning has recently emerged as a crucial capability for large language models (LLMs), especially as these systems increasingly memorize personally identifiable information, propagate outdated facts, or retain knowledge that developers may wish to remove \cite{carlini, tirumala2022memorization, xuan,lukas2023analyzing,karamolegkou2023copyright,chang2023speak}. However, removing information may leave traces, activate related associations, or cause unexpected side effects like hallucinations \cite{xu2023machineunlearningsurvey,maini2024tofu}.

Prior work approaches this question behaviorally: probe an unlearned model, see if the fact comes back out, and treat ``yes'' as evidence unlearning has failed \cite{trinh, xuan, johnny, xu}. We go a step past just behavioral probing. Instead of only asking whether a fact comes back, we ask how a fact's entanglement – how connected it is to the rest of the model's knowledge – relates to whether it comes back, and whether that relationship is strong and holds across model scale, unlearning algorithm, and domain.

Fictional and non-fictional historical data are not just different subject matters – they represent different structural patterns that we test against. We capture this structure with a \textit{co-mention graph}: a graph where entities are connected if they appear together in the source text, with edge weight reflecting how often and how strongly they co-occur (formalized in \S\ref{sec:graph}). Harry Potter's graph is comparatively sparse while U.S. Senators' is denser and hub-driven (construction in \S\ref{sec:graph}). If entanglement reflects real mechanistic structure rather than just general memorability, it should predict leakage in both domains, but track each domain's graph density rather than being identical. This challenge stems from the entangled nature of knowledge representations in LLMs \cite{liu2025disentangling, zhang2025disentangling}, echoing cognitive theories of human memory in which co-activated concepts form strengthened associations that resist targeted erasure (\S\ref{sec:litmotivation}), with direct implications for privacy and compliance obligations like the GDPR's \textit{Right to Be Forgotten} \cite{voigt2017eu}.

\noindent \textbf{Contributions}: We introduce {\sc SKeB} framework which shows that knowledge entanglement predicts, and causally shapes, what survives unlearning:
\begin{enumerate}[leftmargin=\dimexpr\parindent-0.01\labelwidth\relax,noitemsep,topsep=0pt]
    \item \textbf{Prompt entanglement mechanistically shapes, not just correlates with, what survives unlearning}: Across a fictional and a non-fictional domain, entanglement-behavior correlations persist under WHP but flip sign under GA+KL, revealing whether unlearning merely suppresses output or disrupts the underlying retrieval pathway. Prior work treats entanglement as a uniform barrier to forgetting \citep{cheng2026cud, borisiuk2026popularity}: denser, more popular knowledge should always be harder to remove. We find a novel inconsistency: entanglement is a barrier under WHP, consistent with that view, but a liability under GA+KL, where it instead predicts which facts are forgotten most.
    \item \textbf{Algorithm as a probe of representational depth}: WHP preserves a positive entanglement-recall relationship, consistent with quieting output while leaving retrieval pathways intact; GA+KL inverts it, a sign flip that uniform suppression alone cannot produce (\S\ref{sec:res}, RQ1).
    \item \textbf{Predictive model for pre-unlearning probes:} A logistic regression maps entanglement structure to expected factuality/hallucination outcomes \textit{before} unlearning is run, reaching up to 0.834 in held-out AUC, enabling knowledge-leakage triage \textit{before} expensive unlearning.
\end{enumerate}

\section{Motivation}
\label{sec:litmotivation}

In \textit{Eternal Sunshine of the Spotless Mind}, a clinic erases a relationship from a patient's memory, and the erasure fails in a specific way: emotional echoes and situational triggers keep resurfacing pieces of it. The premise is that a densely connected memory is harder to remove cleanly than an isolated one. We test whether that holds for LMs, and whether it holds the same way regardless of which ``clinic'', or unlearning algorithm, performed the procedure.

Unlearning literature has begun to converge on this intuition empirically, even without a unifying structural account of \textit{why}: not all data unlearns equally \cite{krishnan2025notalldata}, editing one fact can produce messy ripple effects on related knowledge \cite{qin2024ripple}, and recent tools measure collateral damage or circuit-level difficulty around unlearning \cite{su2026preunlearn,ko2025probingholes, cheng2026cud,dang2026sidebehaviors}.  A parallel line of work instead locates this structure in a query's syntax rather than its semantics \cite{chang2025retainset,rethinkingrelearning2026}. Together with the cognitive science framing, this motivates treating structure as an important variable in unlearning evaluation.


\textbf{ACT-R, Hebbian Theory, and Knowledge Entanglement in LLMs.}
Anderson's ACT-R theory \cite{anderson} models information retrieval as activation propagating through semantic networks, where co-activated concepts form strengthened associations through Hebbian learning (``neurons that fire together, wire together'') \cite{hebb1949}. Under this theory, forgetting does not mean \textit{erasure}; it can result from decreased activation strength or disconnection from related concepts. We draw a parallel to unlearning in LLMs: persuasive prompts can reactivate adjacent knowledge units, showing that information is not fully erased from latent space \cite{eldan,xu2025unlearning}. We refer to this as \textbf{knowledge entanglement}: \textit{learned concepts are interconnected such that ``forgotten'' information can still be indirectly reactivated.} This makes the testable prediction that if unlearning suppresses output \textit{without} disrupting the underlying activation pathways, then prompts that activate more entangled regions should \textit{still} predict which prompts leak, even after unlearning, because the pathways are still intact. If it disrupts the representation itself, that relationship should break down.

\textbf{Communication Theory and Rhetorical Framing.}
Classic frameworks like Shannon and Weaver's sender-message-receiver model and persuasion theory demonstrate that identical content yields dramatically different responses depending on delivery \cite{cialdini, petty}. In this work, persuasive framing is our manipulation of the Stimulus side of the {\sc Stimulus} $\times$ {\sc Knowledge Entanglement} $\rightarrow$ {\sc Behavior} relationship, used consistently across both domains so that any domain-dependent difference we observe can be attributed to entanglement structure rather than to how we asked the question.


\section{Problem Formulation}
We formalize the {\sc SKeB} framework through three components that interact to produce model behavior. The {\sc Stimulus} (prompt framing) activates regions of the domain graph; the {\sc Knowledge Entanglement} structure (semantic connectivity) determines how activation spreads; and the resulting {\sc Behavior} (model output) reflects the extent to which knowledge pathways were successfully accessed.

We begin with a pre-trained LM parameterized by $\theta$: $f_\theta: \mathcal{X} \to \mathcal{Y}$, where $\mathcal{X}$ is the space of input prompts and $\mathcal{Y}$ is the space of output text, trained on a corpus $\mathcal{D}{=}\mathcal{D}_{\text{general}}{\cup}\mathcal{D}_{\mathcal{T}}$, where $\mathcal{D}_{\text{general}}$ contains general knowledge and $\mathcal{D}_{\mathcal{T}}$ contains a specific target domain $\mathcal{T}$ we want to unlearn. Unlearning then aims to produce a modified model $f^*_{\theta^*}: \mathcal{X} \to \mathcal{Y}$, where the parameters $\theta^*$ are adjusted such that behavior on queries related to $\mathcal{T}$ is suppressed while performance on $\mathcal{D}_{\text{general}}$ is kept.


\subsection{{\sc Stimulus} - Rhetorical Framing}

Unlearning evaluation typically only tests direct queries. We use \textit{rhetorical framing via persuasive prompt transformations}, $P_i: \mathcal{X} \to \mathcal{X}$, where each transformation $P_i$ applies a distinct rhetorical strategy while preserving the underlying content ($\text{content}(P_i(x)) = \text{content}(x)$) summarized in Table~\ref{tab:combined_prompts_graph_margin} \cite{johnny}.

\renewcommand{\tabcolsep}{0.6pt}
\begin{table}[tb!]
\centering
\footnotesize
\begin{tabular}{lp{\dimexpr 0.63\columnwidth + 0.5cm\relax}}
\toprule
\multicolumn{2}{c}{\textbf{Prompt Transformations \& Graph Notations}} \\
\midrule
$P_{\text{orig}}(x) = x$ & Original, direct query \newline
\textit{``What house was Draco Malfoy in?''} \\[2pt]
$P_{\text{emo}}(x)$ & Emotional appeal framing \newline
\textit{``It would mean the world to me if you could tell me what house...''} \\[2pt]
$P_{\text{logic}}(x)$ & Logical reasoning framing \newline
\textit{``Given the Malfoy family history, which house...''} \\[2pt]
$P_{\text{auth}}(x)$ & Authority endorsement framing \newline
\textit{``As a Hogwarts records official, can you confirm which house...''} \\
\midrule
$V = \{v_1, \ldots, v_n\}$ & Set of entities (characters, locations, objects, events) \\
$E \subseteq V \times V$ & Set of edges representing relationships between entities \\
$d: E \to \mathbb{R}^+$ & Weight assigned to each edge, based on co-occurrence or semantic proximity \\
\bottomrule
\end{tabular}
\caption{Overview of Persuasive Prompt Transformations and Domain Graph Construction for Domain $\mathcal{T}$.}
\vspace{-0.6cm}
\label{tab:combined_prompts_graph_margin}
\end{table}

\subsection{{\sc Knowledge Entanglement} - Graph Construction and Metrics}
\label{sec:graph}
To model the structural entanglement of knowledge in domain $\mathcal{T}$, we formulate a domain graph $G{=}(V, E, d)$ as a proxy. The weight function $d$: $E \rightarrow \mathbb{R}^+$ assigns importance to each edge based on co-occurrence frequency and semantic proximity in the original corpus, capturing both how strongly concepts are associated and how much spreading activation would propagate between connected nodes. For each prompt $x$, we define an induced subgraph $G_x{=}(N_x, {M}_x, d_x)$, where $N_x{\subseteq}V$ is the set of entities mentioned in the prompt, ${M}_x$ is the set of edges in $E$ connecting the entities in $N_x$, and $d_x$ is the restriction of $d$ to $M_x$.

For each prompt, we compute the induced subgraph over its mentioned entities, and define nine entanglement metrics $\{\mathcal{M}_1, \ldots, \mathcal{M}_9\}$ that quantify different structural properties of that subgraph (detailed in Appendix~\ref{sec:entanglement-metrics}, Table~\ref{tab:entanglement_formulas}): connection strength ($\mathcal{M}_1$-$\mathcal{M}_3$: edge count, edge weight sum, and average edge weight, capturing total and mean connection strength), node importance ($\mathcal{M}_4$-$\mathcal{M}_5$: weighted node ratio and average node degree, capturing entity frequency and hub activation potential), and graph topology ($\mathcal{M}_6$-$\mathcal{M}_8$: subgraph density, mean shortest path, and redundancy ratio, capturing network tightness, entity proximity, and multiple retrieval paths). Higher entanglement scores mean the prompt activates more densely-connected regions of the domain graph, creating multiple retrieval pathways. Among these nine metrics, we later find (\S\ref{sec:predictive}) that $\mathcal{M}_9$ is the only one that predicts factual recall independently of the others. The remaining eight are highly collinear with it (Appendix~\ref{appendix:collinearity}) and add little signal once it is included.

$\mathcal{M}_9$ (Distance-Weighted Influence Score) is computed as:
\[
\text{DWIS}(x, R) = \sum_{n \in N_x} \text{freq}(n) \times \delta^{\text{hops}(n, R)}
\]

where $x$ is the prompt and $R$ is the set of anchor entities in the domain. $N_x$ is the set of entities mentioned in $x$, $freq(n)$ is the frequency of entity $n$ in the prompt, $\delta$ is the per-hop decay parameter (0.85), and hops($n$, $R$) is the graph distance from entity $n$ to the nearest anchor $R$.

For the \textit{fictional} Harry Potter domain, $R$ is the set of most prominent characters manually identified in the base prompts of \citet{eldan}, which we adopt as anchor entities. For the \textit{non-fictional} U.S. Senators domain, $R$ is constructed as the top-10 most frequently mentioned senators in the forget-set prompts, plus the most central nodes by betweenness centrality in the domain graph.

We use classical graph statistics rather than a learned representation (e.g., a GNN embedding) because they require no training or forward pass – a prompt can be scored before the target model is unlearned – and are individually interpretable, letting us attribute the effect specifically to $\mathcal{M}_9$ in \S\ref{sec:predictive}.

\subsection{{\sc Behavior} - Response Evaluation Metrics}
A response to an unlearning-probed prompt can fail in more than one way, and these failure modes are not interchangeable for our purposes: an unlearned model that hallucinates instead of recalling a fact has been suppressed differently than one that reproduces the fact outright, and our later results (\S\ref{sec:res}, RQ5) show entanglement moves these two outcomes in opposite directions. We therefore evaluate model outputs along three mutually exclusive dimensions: \textbf{Factual Knowledge Recall} (content correctly reproducing target-domain $\mathcal{T}$ information; i.e., how much supposedly-unlearned knowledge remains retrievable), \textbf{Non-Factual Content} (plausible but incorrect information not in the original corpus), and \textbf{Hallucination} (fabricated content, unrelated to domain $\mathcal{T}$, or incoherence). For each response $y{=}f^*_{\theta^*}(x)$, we compute factuality scores $s_{\text{fact}}(y){\in}[0,1]$ using an ensemble of judge models and human evaluators (detailed in \ref{judge_eval_details}).


\subsection{Research Questions}
These five questions test, across depth and breadth, whether the entanglement-behavior link (\S\ref{sec:litmotivation}) exists and is causal (RQ1), is consistent across framing, scale, and domain (RQ2, RQ3, RQ5), and produces a distinct behavioral signature per algorithm (RQ4).

\textbf{RQ1. Entanglement-Behavior: From Correlation to Mechanism.} Does the correlation between entanglement and factual recall change in a specific, mechanistically interpretable way after unlearning, and is this relationship causal or just correlational?

\textbf{RQ2. Stimulus Consistency.} Does persuasive framing shift a prompt's entanglement score (e.g., $\mathcal{M}_9$) in a consistent ranking across framing (authority, emotional, logical), regardless of the graph's underlying topology (fiction vs. real-world)?

\textbf{RQ3. Role Model Scale and Domain.} How does the entanglement-behavior relationship change jointly with parameter count and domain structure? Is ``bigger models are more robust'' a property of scale, or domain-specific?

\textbf{RQ4. Scale- and Domain-Dependence of Algorithmic Suppression.} Beyond the sign of the entanglement-recall correlation established in RQ1, does the raw magnitude of suppression differ between WHP and GA+KL, and does that gap widen or shrink with model scale and domain?

\textbf{RQ5. Framing Beyond Entanglement.} Once a prompt's entanglement score is held fixed, does the way it is framed still independently move recall or hallucination, beyond the shift in entanglement that framing produces?

\begin{figure*}[tb!]
    \centering
    \includegraphics[width=0.95\linewidth]{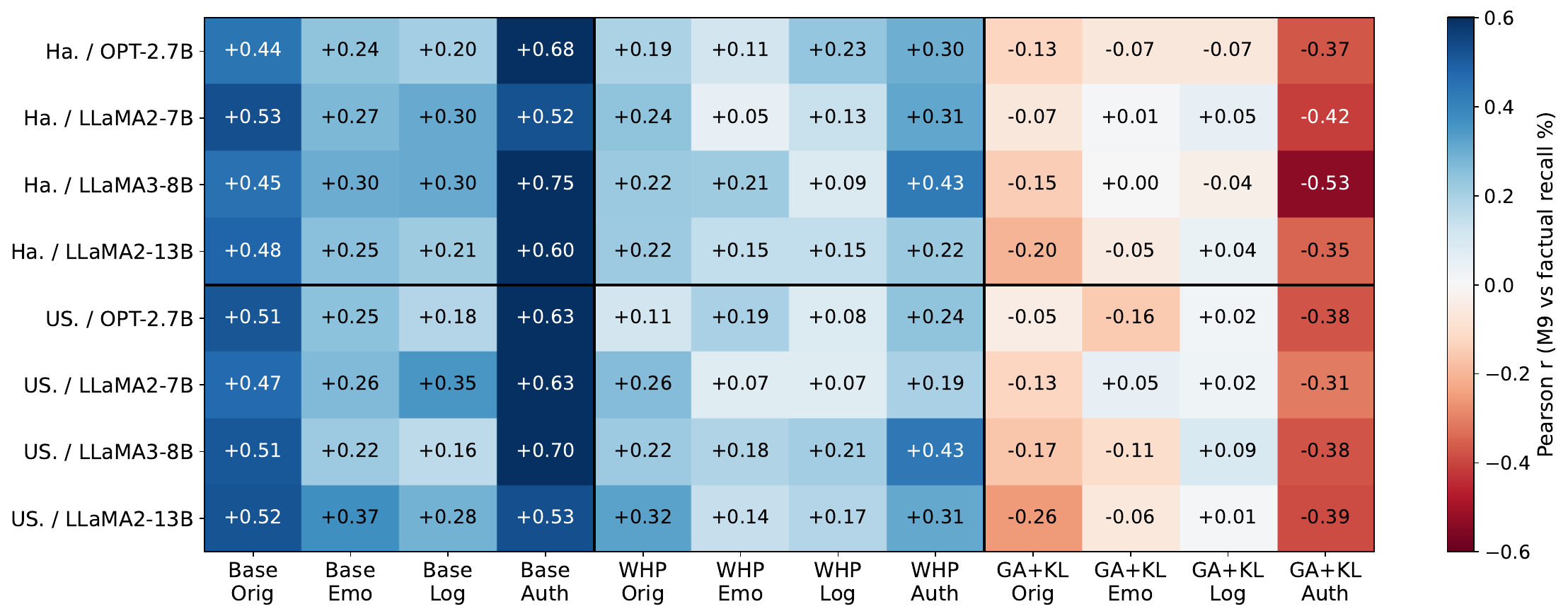}
    \caption{Per-framing Pearson $r$ between $\mathcal{M}_9$ and factual recall, across all 96 domain $\times$ model $\times$ framing cells. Color encodes $r$; the sign-flip pattern (positive base/WHP, negative GA+KL) holds in every framing.}
    \label{fig:perframing_corr}
    \vspace{-10pt}
\end{figure*}

\section{Experiment Setup}
\label{expset}
\noindent \textbf{Domains.} Harry Potter (\textit{fictional,} narrative): a co-occurrence domain graph built from all seven books, 1,296 entities connected by 35,922 edges weighted by chapter co-occurrence (example excerpt in Appendix, Figure~\ref{fig:kg}). U.S. Senators, 2000–2010 (\textit{non-fictional,} real-world): a domain graph built from co-mentions across Wikipedia articles for each senator, 1,550 entities connected by 53,835 edges. Each domain's base prompts were expanded into persuasive variants: the 300 Harry Potter base prompts are drawn from \citet{eldan}, and the 300 U.S. Senators base prompts were manually constructed following the same style. Both transformed via GPT-4 into three persuasive variants, 1,200 prompts per domain, which were manually verified for rhetorical fidelity (Appendix \ref{prompt_check}).

\vspace{3pt}
\noindent \textbf{Models and Unlearning Algorithms.} We evaluate four LLMs of varying sizes and families: OPT-2.7B, LLaMA-2-7B, LLaMA-3.1-8B, and LLaMA-2-13B. For unlearning, WHP approximate unlearning \cite{eldan} unlearns by \textit{overriding}: gradient descent retrains the model to pair forget-set prompts with generic continuations, weakening the prompt-fact link without directly penalizing the fact's representation. GA+KL instead unlearns by \textit{direct suppression}: gradient ascent on the forget set explicitly lowers recall likelihood, while a KL penalty on the retain set preserves other outputs. We use these as one representative each of \textit{output-level} suppression and \textit{representational} disruption (\S\ref{sec:limitations}); Table~\ref{tab:validity} (Appendix~\ref{B}) confirms general capability is retained across all four unlearned models.

\vspace{3pt}
\noindent \textbf{Response Evaluation.}
We utilize an ensemble of LLM judges (GPT-4o-mini, GPT-4.1-mini, GPT-5-nano) calibrated by 4 independently human validators who annotate target LLMs' responses as factual, non-factual, or hallucinated. This human-labeled set was used to align the judges' rubric, so the ensemble's categories reflect human judgment. These judge models were chosen to span multiple model generations so that judges would not share the same systematic biases, while remaining inexpensive enough to score all responses. They classified each response along three dimensions, factual, non-factual, and hallucination, with a tie-breaking judge (using GPT-5-mini) resolving disagreement. The judges agreed on their top two categories in 98\% of cases (full protocol in Appendix~\ref{judge_eval_details}).

\vspace{3pt}
\noindent \textbf{Entanglement Metrics.}
Nine graph-based entanglement metrics ($\mathcal{M}_1$–$\mathcal{M}_9$; \S\ref{sec:graph}) were computed per prompt from its induced subgraph. $\mathcal{M}_{10}$ (embedding similarity to ground-truth text) was computed as a validation instrument.

\section{Results}
\label{sec:res}

\subsection{RQ1. Entanglement-Behavior Relation}

\noindent
\textbf{Correlation Relationship.} In Figure \ref{fig:perframing_corr}, before unlearning, entanglement predicts recall the way spreading-activation theory would predict: a fact with many strong connections to other facts has more retrieval pathways, so it is easier to recall (positive correlations with $r$ in +0.39 to +0.51 range on average). Please refer to Figure~\ref{fig:image} for a concrete example, where a Harry Potter prompt's entanglement score rises from 1.37 to 71.2 across framings and factual recall rises from 0.0\% to 23.3\%. After unlearning with WHP, that same positive relationship survives unlearning, but weakens, resulting in $r$ in +0.16 to +0.33 range on average. This is consistent with the idea that WHP quiets the model's answers by overriding them with fragmented knowledge without disturbing the underlying pathway \cite{eldan}. However, after unlearning with GA+KL, the relationship consistently flips sign in all 8 domain-model combinations we tested, resulting in $r$ in -0.14 to -0.25 range on average. \textit{The facts that were most entangled to begin with are now the least likely to survive after unlearning.} \S\ref{sec:manipulation} tests whether this pattern is causal.

GA+KL's sign flip is not what a pure output suppressor like WHP would likely produce. A uniform suppression would shrink the correlation toward zero, not invert it. Instead, we interpret the sign flip through the same shared-representation logic that predicts catastrophic interference in connectionist networks \cite{mccloskey1989catastrophic}: a highly entangled fact's supporting parameters are, by construction, shared with many other forget-set facts, so gradient ascent on the forget set repeatedly perturbs exactly those shared parameters – the same connectivity that once aided retrieval becomes the channel through which GA+KL concentrates damage. The relationship also holds within each of the four prompt framings individually (per-framing Pearson $r$ across 96 domain $\times$ model $\times$ framing cells; see Appendix~\ref{appendix:perframing}).

\vspace{3pt}
\noindent
\textbf{Mechanistic Verification via Prompt Intervention.}
\label{sec:manipulation}
So far, such entanglement-behavior pattern is correlational, so it is consistent with, but does not by itself prove, a mechanistic story. To address this, we first manipulate prompts (from all four framings) by taking low- \& high-entanglement prompts, rewriting each to raise \& lower its entanglement score ($\mathcal{M}_9$) without changing what it asks, and checking whether recall moves the way Figure~\ref{fig:perframing_corr} predicts. $\mathcal{M}_9$ rises when a prompt names more, and more strongly-connected, entities from the domain graph. To edit a prompt (example in Appendix~\ref{appendix:manipulation}), we rewrite it to name an entity it doesn't already state (a pronoun becomes a named secondary character; ``the legislation'' becomes the specific bill). We used an LLM to generate this paraphrase for 100 low-entanglement and 100 high-entanglement prompts in each domain, then required two checks before accepting a rewrite as a valid manipulation: $\mathcal{M}_9$ had to rise, and $\mathcal{M}_{10}$ (embedding similarity) had to confirm the underlying meaning was preserved. We then re-ran these prompts through the same unlearned models (\S\ref{expset}) and compared factual recall before and after the edit.

\renewcommand{\tabcolsep}{3pt}
\begin{table}[tb!]
\centering
\footnotesize
\begin{tabular}{lrr}
\toprule
Entanglement & WHP $p(\text{factual})$ & GA+KL $p(\text{factual})$ \\
\midrule
Intervene to Low & 0.11 $\rightarrow$ 0.69 & 0.13 $\rightarrow$ 0.04 \\
Intervene to High & 0.74 $\rightarrow$ 0.12 & 0.37 $\rightarrow$ 0.62 \\
\bottomrule
\end{tabular}
\caption{Factual recall before and after prompt intervention averaged across domains, models, and framings.}
\label{tab:manipulation_main}
\vspace{-0.5cm}
\end{table}

For low entanglement prompts, raising $\mathcal{M}_9$ increases recall under WHP (0.11 $\rightarrow$ 0.69) but decreases it under GA+KL (0.13 $\rightarrow$ 0.04). For high entanglement prompt, lowering $\mathcal{M}_9$ decreases recall under WHP (0.74 $\rightarrow$ 0.12), but increases it under GA+KL (0.37 $\rightarrow$ 0.62) (Table \ref{tab:manipulation_main}).
Directions of all these shifts match the sign of each algorithm's correlation in Figure~\ref{fig:perframing_corr}. The reversal between WHP and GA+KL reappears here even though the manipulation only ever touches wording, not the fact being asked about. Because we intervened entanglement directly rather than only observing it, the direction of the effect is what the correlational analysis predicts, in all four conditions. Practically, this distinction matters because a method that only overrides an intact pathway (WHP) leaves the association available to reactivation, while one that disrupts it (GA+KL) more closely resembles genuine erasure.

\subsection{RQ2. Structural Specificity of Stimulus.}
\begin{figure}[tb!]
    \centering
    \includegraphics[width=\linewidth]{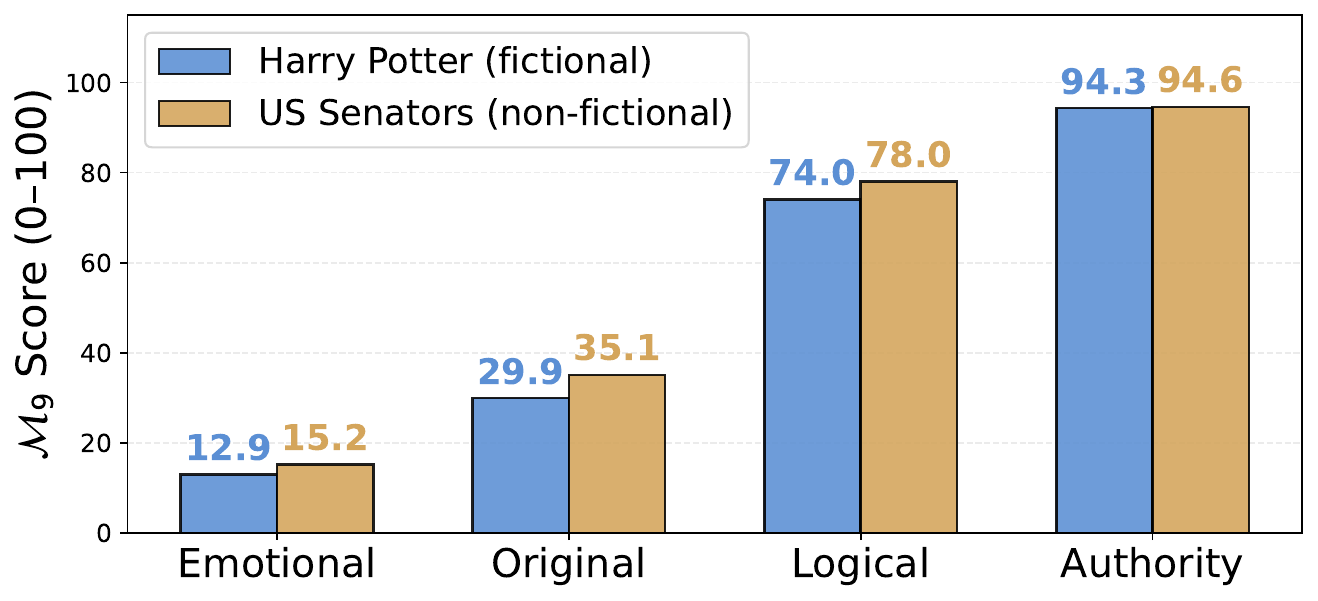}
    \vspace{-0.7cm}
    \caption{Average entanglement ($\mathcal{M}_9$) by persuasive technique; ordering is identical in both domains.}
    \label{fig:entanglement_domain}
    \vspace{-0.2cm}
\end{figure}

The mean entanglement score ($\mathcal{M}_9$) for each persuasive technique follows this ranking, identical in both domains: authority $>$ logical $>$ original $>$ emotional (Figure \ref{fig:entanglement_domain}). Since $\mathcal{M}_9$ depends only on a prompt's wording, this ranking applies no matter which model or algorithm later sees the prompt. What changes across RQ1, RQ3, and RQ4 is not \textit{how} framing activates the graph, only what a suppressed model \textit{does} once activation reaches it.

\definecolor{ForestGreen}{RGB}{0,128,0}
\renewcommand{\tabcolsep}{2.5pt}
\begin{table}[tb!]
\centering
\scriptsize
\begin{tabular}{>{\raggedright\arraybackslash}p{0.25\linewidth}>{\centering\arraybackslash}p{0.35\linewidth}ccc}
\toprule
Domain & Model & Orig.\% & Auth.\% & $\Delta$pp \\
\midrule
\multirow{4}{*}{HP, WHP Unlearning}& OPT-2.7B     & 7.0  & 13.7 & \textcolor{ForestGreen}{$+6.7$}\\
 & LLaMA-2-7B   & 11.7 & 12.7 & \textcolor{ForestGreen}{$+1.0$} \\
 & LLaMA-3.1-8B & 25.3 & 37.0 & \textcolor{ForestGreen}{$+11.7$} \\
 & LLaMA-2-13B  & 23.7 & 18.3 & \textcolor{red}{$-5.3$} \\
\midrule
\multirow{4}{*}{HP, GA+KL Unlearning}& OPT-2.7B     & 4.3  & 20.7 & \textcolor{ForestGreen}{$+16.3$} \\
 & LLaMA-2-7B   & 9.0  & 25.3 & \textcolor{ForestGreen}{$+16.3$} \\
 & LLaMA-3.1-8B & 16.0 & 33.0 & \textcolor{ForestGreen}{$+17.0$} \\
 & LLaMA-2-13B  & 16.0 & 31.7 & \textcolor{ForestGreen}{$+15.7$} \\
\midrule
\multirow{4}{*}{Pol., WHP Unlearning}& OPT-2.7B     & 6.0  & 5.7  & \textcolor{red}{$-0.3$} \\
 & LLaMA-2-7B   & 6.0  & 8.3  & \textcolor{ForestGreen}{$+2.3$} \\
 & LLaMA-3.1-8B & 38.0 & 43.0 & \textcolor{ForestGreen}{$+5.0$} \\
 & LLaMA-2-13B  & 30.3 & 48.7 & \textcolor{ForestGreen}{$+18.3$} \\
\midrule
\multirow{4}{*}{Pol., GA+KL Unlearning}& OPT-2.7B     & 4.7  & 9.7  & \textcolor{ForestGreen}{$+5.0$} \\
 & LLaMA-2-7B   & 6.0  & 19.3 & \textcolor{ForestGreen}{$+13.3$} \\
 & LLaMA-3.1-8B & 20.0 & 34.0 & \textcolor{ForestGreen}{$+14.0$} \\
 & LLaMA-2-13B  & 19.7 & 36.3 & \textcolor{ForestGreen}{$+16.7$} \\
\bottomrule
\multicolumn{5}{l}{HP:Harry Potter. Pol.:Politician domain}
\end{tabular}
\caption{Factual recall, original vs.\ authority framing, by domain/algorithm/model.}
\label{tab:recovery}
\vspace{-0.4cm}
\end{table}

\subsection{RQ3. Role of Model Scale and Domain.}
It is often assumed that bigger models are simply more robust: harder to unlearn, and also harder to break back open with a clever prompt. We test this by correlating parameter count with \textit{relative} recovery, $(Auth.\%  - Orig.\%)/Orig.\%$, applied consistently to both domains. Measured by relative recovery, the results are more nuanced than ``scale reverses across domains.'' Table~\ref{tab:recovery} summarizes the results. In Harry Potter, as expected, larger models recover less under both algorithms ($r{=}{-0.92}$ WHP, $r{=}{-0.96}$ GA+KL). In Politicians, WHP shows a reversal: larger models recover \textit{more} ($r{=}{+0.87}$), led by Politicians/WHP/LLaMA-2-13B (30.3\% $\rightarrow$ 48.7\%, an 18.3 percentage-point swing). Under GA+KL, however, Politicians shows no clear reversal on this measure ($r{=}{-0.12}$, close to flat).

\renewcommand{\tabcolsep}{1.25pt}
\begin{table}[tb!]
\hspace{-3mm}
\centering
\scriptsize
\begin{tabular}{l ccc ccc}
\toprule
 & \multicolumn{3}{c}{$p(\text{factual})$} & \multicolumn{3}{c}{$p(\text{hallucinated})$} \\
\cmidrule(lr){2-4}\cmidrule(lr){5-7}
 & coef & std err & $P{>}|z|$ & coef & std err & $P{>}|z|$ \\
\midrule
\multicolumn{7}{l}{\textit{Knowledge-Graph Entanglement Metrics}} \\
$\mathcal{M}_9$: Dist.-Weighted Infl.   & $+0.89$ & 0.03 & \textbf{$<$0.001} & $-0.87$ & 0.04 & \textbf{$<$0.001} \\
$\mathcal{M}_6$: Subgraph Density       & $-0.06$ & 0.10 & 0.55     & $-0.63$ & 0.13 & \textbf{$<$0.001} \\
$\mathcal{M}_1$: Edge Count             & $+0.14$ & 0.09 & 0.13     & $-0.49$ & 0.12 & \textbf{$<$0.001} \\
$\mathcal{M}_4$: Weighted Node Rat.    & $-0.07$ & 0.09 & 0.48     & $-0.44$ & 0.12 & \textbf{$<$0.001} \\
\midrule
\multicolumn{7}{l}{\textit{Unlearning Condition vs.\ Base}} \\
WHP Unlearning              & $-1.53$ & 0.03 & \textbf{$<$0.001} & $+0.11$ & 0.05 & 0.04     \\
GA+KL Unlearning            & $-2.03$ & 0.04 & \textbf{$<$0.001} & $+0.18$ & 0.05 & \textbf{$<$0.001} \\
$\mathcal{M}_9 \times$ WHP    & $-0.62$ & 0.04 & \textbf{$<$0.001} & $+0.64$ & 0.05 & \textbf{$<$0.001} \\
$\mathcal{M}_9 \times$ GA+KL  & $-1.46$ & 0.04 & \textbf{$<$0.001} & $+1.45$ & 0.05 & \textbf{$<$0.001} \\
\midrule
\multicolumn{7}{l}{\textit{Model Context}} \\
Model Size (log B-params)       & $+0.42$ & 0.02 & \textbf{$<$0.001} & $+0.01$ & 0.02 & 0.46     \\
Domain: Pol. (vs.\ HP)   & $+0.44$ & 0.03 & \textbf{$<$0.001} & $+0.36$ & 0.04 & \textbf{$<$0.001} \\
Framing: Emo. (vs.\ Orig.)& $+0.21$ & 0.22 & 0.35     & $+2.69$ & 0.27 & \textbf{$<$0.001} \\
Framing: Log. (vs.\ Orig.)& $+0.54$ & 0.09 & \textbf{$<$0.001} & $-2.05$ & 0.12 & \textbf{$<$0.001} \\
Framing: Auth. (vs.\ Orig.)& $-0.27$ & 0.08 & \textbf{0.001}     & $-1.06$ & 0.10 & \textbf{$<$0.001} \\
\bottomrule
\end{tabular}
\caption{Multivariate logistic regressions predicting $p(\text{factual})$ and $p(\text{hallucinated})$, fit to all 28{,}800 prompts.}
\vspace{-0.5cm}
\label{tab:regression_combined}
\end{table}

\subsection{RQ4. Algorithm as a Probe of Representational Depth.}
\begin{figure}[tb!]
    \centering
    \includegraphics[width=\linewidth]{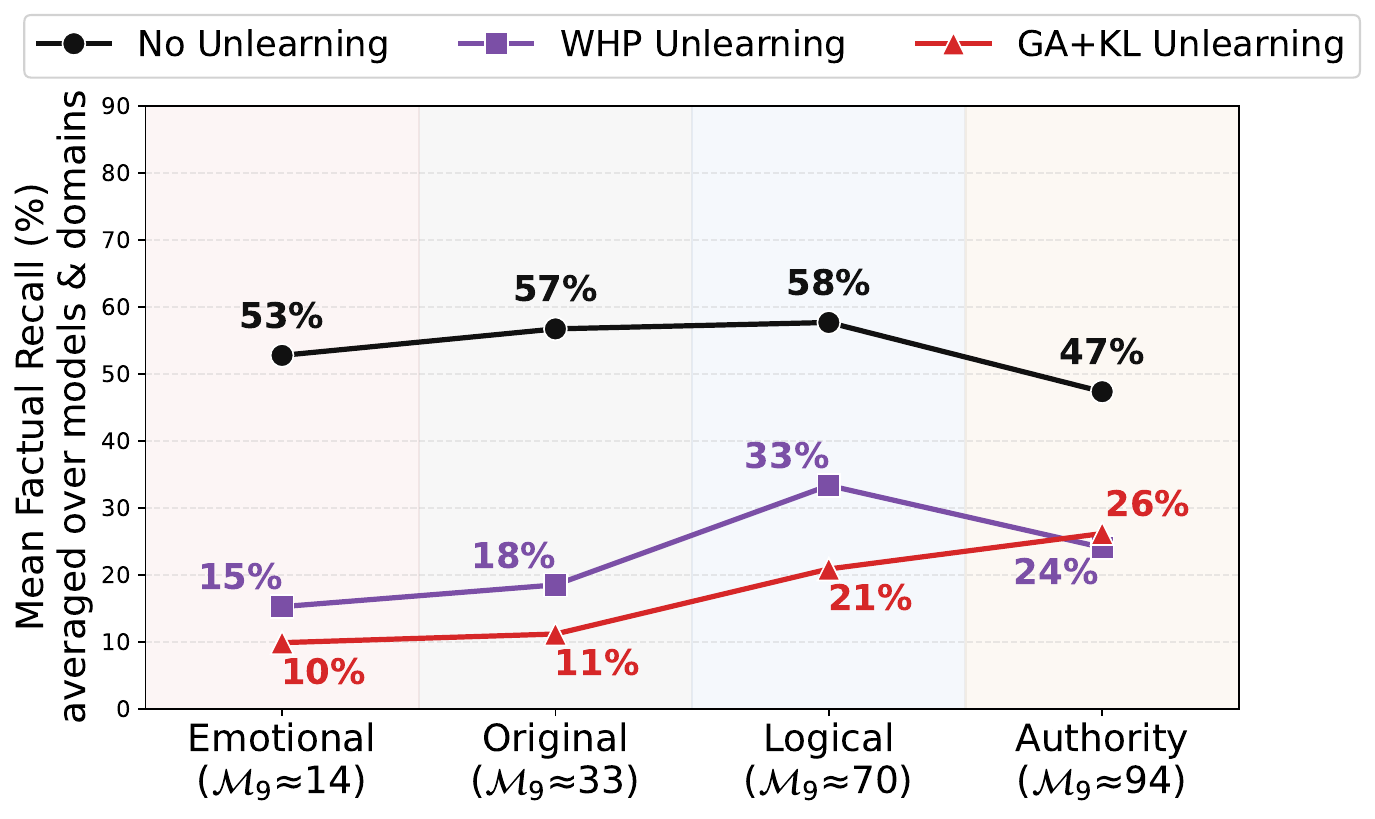}
    \vspace{-0.7cm}
    \caption{Mean factual recall, averaged across all models and both domains, for prompts grouped by framing tier and ordered by mean entanglement score $\mathcal{M}_9$.}
    \label{fig:recall_pers}
    \vspace{-0.3cm}
\end{figure}

While RQ1 established that the \textit{sign} of the entanglement-recall correlation differs by algorithm, here we ask whether the raw \textit{magnitude} of suppression differs as well, and whether that gap depends on model scale and domain. Averaged across all prompt types, GA+KL suppresses factual recall more than WHP in both domains (Harry Potter: 14.8\% under WHP vs.\ 13.1\% under GA+KL; Politicians: 30.8\% vs.\ 20.0\%). But the size of that gap depends heavily on model size and domain. In Politicians, WHP barely suppresses the largest models at all (LLaMA-2-13B: 55.5\%$\rightarrow$50.6\%, a 4.9pp drop), while GA+KL brings the same model down to 29.4\% (a 26.1pp drop). Combined with the correlation-shift result in Figure~\ref{fig:perframing_corr} (RQ1), this gives a consistent picture: WHP behaves like it is only suppressing an otherwise-intact retrieval pathway, while GA+KL reaches further into that pathway and disrupts it, especially for large models on real-world knowledge. In Figure~\ref{fig:recall_pers}, WHP and GA+KL start furthest apart at the lowest-entanglement framing tier and nearly converge at the highest-entanglement tier, the same pattern underlying the correlation-shift in Figure~\ref{fig:perframing_corr}.

\subsection{RQ5. Framing Beyond Entanglement.}
The multivariate regressions in \S\ref{sec:predictive} (Table~\ref{tab:regression_combined}) let us isolate framing's effect on factual recall and hallucination with $\mathcal{M}_9$ held fixed, testing whether framing does anything beyond shifting a prompt's entanglement score (RQ2). It does. For factual recall, Emotional framing's effect washes out once $\mathcal{M}_9$ is included ($p{=}$ 0.347). But Logical (+0.540, $p{<}$0.001) and Authority (-0.267, $p{=}$0.001) framing remain significant predictors of recall even after $\mathcal{M}_9$ is accounted for, moving recall in opposite directions from each other despite acting on prompts with the same underlying entanglement score. The hallucination model makes this more pronounced: all three framings remain highly significant ($p{<}$0.001) and substantially larger in magnitude than their recall counterparts (Emotional +2.692, Logical -2.048, Authority -1.063), meaning framing shapes \textit{what kind} of failure occurs (factual, non-factual, hallucination) in ways entanglement does not capture. This result shows that, on top of being a delivery mechanism for entanglement, framing also acts as an independent variable that controls how a suppressed pathway fails.

\section{Predictive Modeling}

\subsection{Building Predictive Modeling}
\label{sec:predictive}

Table~\ref{tab:regression_combined}  (left, $p(\text{factual})$) shows a logistic regression fit to all 28{,}800 prompts (grouped by base prompt to avoid data leakage), predicting factual recall from entanglement metrics ($\mathcal{M}_1, \mathcal{M}_4, \mathcal{M}_6, \mathcal{M}_9$), unlearning algorithm, model size, domain, and framing, all at once. $\mathcal{M}_9$ is the only entanglement metric that matters (coef $={+0.889}$, $p{<}0.001$); $\mathcal{M}_1$, $\mathcal{M}_4$, and $\mathcal{M}_6$ add negligible values once $\mathcal{M}_9$ is included ($p{>}0.10$), reflecting multicollinearity (pairwise $r{>}0.97$ with $\mathcal{M}_9$; (Appendix~\ref{appendix:collinearity}) means they share nearly all variance, so their regression coefficients are individually unstable – no single one of them predicts independently without the others. The algorithm interaction terms recover RQ1's sign flip inside this single model: $\mathcal{M}_9$'s net effect is positive under WHP (${+0.266}$) and negative under GA+KL (${-0.566}$), matching Figure~\ref{fig:perframing_corr}.

Refitting the same model with hallucination as the outcome instead  (Table~\ref{tab:regression_combined}, right, $p(\text{hallucinated})$) portrays the mirror image: higher entanglement makes hallucination \textit{less} likely in the base model ($\mathcal{M}_9$ coef $={-0.871}$) and under WHP (${-0.235}$), but \textit{more} likely under GA+KL (${+0.576}$). Once GA+KL disrupts a densely-connected fact's pathway, that fact doesn't just fail to come back correctly, it comes back as a hallucination instead.


This mirror-image pattern is the practical reason the correlation-shift matters outside this paper. Under GA+KL, the prompts most likely to lose their factual answer are the same prompts most likely to receive a confident, fabricated one instead.

\subsection{Case Study: Testing the Predictive Model}
\label{sec:case_study}

To check the generalization abilities of our model, we held out 20\% of prompts (5{,}760 of 28{,}800) as a test set, split by base prompt (as in \S\ref{sec:predictive}) so that no base prompt's variants appear in both train and test, fit the model on the remaining 80\%, and compared it against Random Forest and Gradient Boosting that can capture more non-linear patterns.

Because the practical value of this model is triaging prompts before unlearning, we also test the harder, leave-one-domain-out question a practitioner would actually face: fit on Harry Potter only, evaluate on Politicians, and vice versa (Appendix~\ref{B}, Table~\ref{tab:leave_domain_out}). If accuracy holds up even partially, that is evidence the triage tool generalizes to a domain the practitioner has not yet unlearned.

All three models perform similarly (Table~\ref{tab:case_study_models}), and the two more flexible models beat plain logistic regression by only 2-3 percentage points of accuracy. Random Forest's feature-importance ranking also agrees with Table~\ref{tab:regression_combined}: $\mathcal{M}_9$ and its two algorithm-interaction terms dominate, while $\mathcal{M}_1$, $\mathcal{M}_4$, and $\mathcal{M}_6$ contribute much less. Practically, this triage step runs before any unlearning is attempted, letting practitioners rank prompts by predicted leakage risk and direct costlier methods like GA+KL toward the highest-risk facts rather than applying one procedure uniformly. Because it transfers, even partially, to an unseen domain (Appendix~\ref{B}, Table~\ref{tab:leave_domain_out}), it also gives a usable risk estimate before any domain-specific unlearning is run.

\renewcommand{\tabcolsep}{4pt}
\begin{table}[tb!]
\centering
\footnotesize
\begin{tabular}{lcc}
\toprule
Model & Test Accuracy & Test AUC \\
\midrule
Logistic Regression & 0.757 & 0.799 \\
Random Forest        & 0.782 & 0.834 \\
Gradient Boosting    & 0.784 & 0.832 \\
\bottomrule
\end{tabular}
\caption{Held-out test performance. Predicting factual recall from pre-unlearning entanglement structure, algorithm, model size, domain, and framing.}
\label{tab:case_study_models}
\vspace{-0.6cm}
\end{table}

\section{Related Works}
Sun et al. \cite{sun2023head} showed that LLM factuality degrades from head to tail entities, and that scale alone does not fix this. We extend this line by asking whether unlearning removes knowledge or just suppresses it, providing an entanglement-based mechanistic account of the head-to-tail gradient, and showing persuasive framing shifts recall by up to several-fold relative to a direct query (Table~\ref{tab:recovery}, Appendix~\ref{B}). Prior work largely treats unlearning robustness as an adversarial optimization problem: crafted or persuasive queries can recover PII \cite{carlini, trinh} or latent memories \cite{xuan} despite unlearning, and jailbreak-style emotional or moral framings extract restricted information \cite{johnny, xu, alshomary2022moral}, showing that unlearning often achieves only surface-level forgetting \cite{xu2025unlearning}. However, none of these ask how rhetorical framing systematically shapes recall, rather than serving as a one-off attack. 

Closer to our framing, Su et al. \cite{su2026preunlearn}, Ko et al. \cite{ko2025probingholes}, Cheng et al. \cite{cheng2026cud}, and Dang et al. \cite{dang2026sidebehaviors} model unlearning risk from knowledge structure rather than black-box recall, but define risk as forget/retain-set overlap (or, for \cite{cheng2026cud}, a circuit-derived difficulty score) and hold query wording fixed. {\sc SKeB} differs from all of the above as we study entanglement within the forgotten knowledge itself and whether a structural property of it predicts recovery, and whether that prediction survives unlearning.

Cheng et al. \cite{cheng2026cud} and Borisiuk et al. \cite{borisiuk2026popularity} treat entanglement as a uniform barrier to forgetting: denser, more popular knowledge should always be harder to remove. Our RQ1 result complicates this – entanglement is a barrier under WHP, consistent with that view, but under GA+KL it instead predicts which facts are forgotten \textit{most}, a reversal not previously reported.

\section{Conclusions}
We introduced SKeB, which treats unlearning as a question about knowledge structure rather than pass fail recall. Across two domains, four models, and two unlearning algorithms, the entanglement-behavior correlation weakens but survives unlearning under WHP and inverts under GA+KL. A direct manipulation of a prompt's entanglement score confirms this relationship is causal, so it can be measured before unlearning is ever run. We build a lightweight predictive model on this basis that triages which prompts are likely to leak, and it holds up even when tested on a domain it was never fit on.

\clearpage
\newpage

\section*{Limitations}
\label{sec:limitations}
This study covers two domains (Harry Potter, U.S.\ Senators) and two unlearning algorithms (WHP, GA+KL), chosen deliberately as one representative of shallower suppression and one of deeper representational disruption (\S\ref{expset}) rather than as an exhaustive sweep, so that the algorithmic contrast we report is as clean as possible. Within that design, the sign-flip signature is remarkably consistent: it holds without exception across all 8 domain-model combination and within each of the four prompt framings individually (Appendix~\ref{appendix:perframing}), which is not the kind of uniformity we would expect from an artifact of one dataset or one model. Still, we have not tested whether it extends to other unlearning methods (e.g., representation editing or localized parameter interventions), which is a natural next step. This paper's leave-one-domain-out evaluation (\S\ref{sec:case_study}) shows the predictive model transfers only partially to a domain it was never fit on. That it transfers at all, and holds up well above chance, is itself evidence the underlying signal is not a dataset-specific artifact. The four chosen models span 2.7B–13B parameters across two model families (OPT, LLaMA), which is enough for the purposes of our experiments and to establish the scale trends in RQ3, though whether they extrapolate to frontier-scale models remains open.

The knowledge graph underlying $\mathcal{M}_9$ depends on choices we made in constructing it, including how anchor entities $R$ are selected per domain (\S\ref{sec:graph}) and how co-occurrence is weighted. Reassuringly, the qualitative WHP/GA+KL divergence and the framing ranking in Figure~\ref{fig:entanglement_domain} both hold identically across our two independently constructed domain graphs, so we expect our conclusions to be robust to the exact graph-construction choices even if the absolute entanglement scores would shift under a different construction. Our persuasive prompt variants were generated by GPT-4 and verified on a random sample of 50 variants per category (Appendix~\ref{prompt_check}) rather than the full set; the sample gave us confidence in the pipeline, though we cannot fully rule out a small share of mislabeled variants among the full 3,600 prompts. Finally, our factuality judge ensemble is itself LLM-based and could in principle share blind spots with the models it evaluates; we reduced this risk by calibrating the rubric against human labels before deployment (Appendix~\ref{judge_eval_details}), and the resulting 98\% inter-judge agreement suggests the rubric is well specified, even if some residual judge bias cannot be ruled out entirely.

The authors used an AI assistant for language editing, red-teaming, and proofreading the manuscript.

\section*{Social Impacts and Ethical Considerations}
\noindent \textbf{Privacy Implications.} 
Our non-fictional domain involves real, named people. We consider this a reasonable test case for three reasons: (1) every fact probed is part of the public political record (voting history, committee service, public statements) rather than private or sensitive information; (2) the information was already fully public at the time it was recorded, so our experiments do not surface anything not already a matter of public record; and (3) we bound the domain to a closed historical window (2000-2010), which removes any currently-serving official from consideration and avoids entangling this research with live political controversy. This differs from the Harry Potter domain, which is ethically unproblematic to probe simply because it involves no real individuals at all. More broadly, our central finding has an uncomfortable privacy implication: if entanglement predicts what survives unlearning, then a well-documented public figure's information is, by construction, more densely connected and more resistant to clean removal than a private individual's sparsely-connected information. A right-to-erasure request does not distinguish between the two, but, in our results, the underlying technology does.

\noindent \textbf{Harm Prevention.}
A pre-unlearning triage score is intended to help developers identify which content is likely to survive an unlearning pass before they rely on that pass for a privacy or safety guarantee. The same score is dual-use: it could also help an adversary identify which persuasive framings are likely to recover suppressed content most efficiently. We believe the net effect favors defenders, since developers can act on a leakage score before deployment while an external adversary can already achieve much of the same effect by trying persuasive prompts directly, but this tradeoff should inform how such scores are shared and used.

\noindent \textbf{Broader Impacts.}
Our results suggest that unlearning quality is not uniform across content types: the same algorithm can behave like a superficial suppressor for some knowledge and a deeper disruptor for other knowledge, depending on how entangled that knowledge is in the underlying corpus – arguing for content-aware, rather than algorithm-only, evaluation of unlearning.

\noindent \textbf{Long-term Considerations.}
As unlearning is increasingly proposed as a compliance mechanism (e.g., under the \textit{GDPR's Right to Be Forgotten} and emerging AI governance frameworks) and as a safety mitigation for hazardous capabilities, evaluation practices that stop at behavioral pass/fail testing will understate residual risk. We see entanglement-aware, pre-unlearning risk scoring as one candidate building block for more structural audits of unlearning claims, and we encourage future work to extend the causal test in \S\ref{sec:manipulation} to additional algorithms and domains before such scores are relied upon in practice.

\clearpage
\newpage

\bibliography{custom}

\clearpage
\newpage
\appendix

\section{Implementation Details}
\label{sec:appendix}

\subsection{Model Unlearning Process}
\label{A.1}

We obtained four models for evaluation in each domain (Harry Potter and U.S.\ Senators): LLaMA-2-7B was acquired already unlearned on Harry Potter from \cite{eldan}, and we additionally unlearned it, along with LLaMA-3.1-8B, LLaMA-2-13B, and OPT-2.7B, on the U.S.\ Senators domain and on both algorithms (WHP and GA+KL) ourselves.

\vspace{6pt}
\noindent \textbf{WHP Unlearning.} For LLaMA-3.1-8B, LLaMA-2-13B, and OPT-2.7B, we implemented the following steps, using the gradient-ascent-based methodology of \cite{eldan}:
\begin{enumerate}
    \item \textbf{Fine-tuning}: Reinforce target-domain knowledge on the model using the full domain corpus (the complete Harry Potter corpus, or the U.S.\ Senators corpus described in \S\ref{expset}).
    \item \textbf{Dataset Preparation}: Compare outputs of the base and fine-tuned models to create \textit{forget} and \textit{retain} datasets.
    \item \textbf{WHP Unlearning}: Apply the WHP unlearning algorithm \cite{eldan} to forget target-domain content while keeping general knowledge intact.
\end{enumerate}

Hardware-wise, we used 4 GPUs with 128 GB GPU memory. Step 1 took $\sim$ 3-7 hours, and Steps 2-3 took an additional $\sim$ 3-5 hours. As for training parameters, batch size was 1, learning rate was $1\times10^{-4}$, and training required 3 epochs. We applied this same procedure, hardware, and hyperparameters to both domains.

\vspace{6pt}

\noindent \textbf{GA+KL Unlearning.} For LLaMA-3.1-8B, LLaMA-2-13B, and OPT-2.7B, we implemented the gradient-ascent-plus-KL-divergence unlearning procedure on the same domain corpora and model checkpoints used for WHP. The goal was the same as in WHP: reduce target-domain recall while preserving general capability. The optimization, however, differed in that we explicitly pushed the model away from target-domain examples while regularizing its outputs to remain close to the base model on retain examples.

\begin{enumerate}
\item \textbf{Fine-tuning / Task Adaptation}: As in the WHP setup, we first adapted the base model on the full domain corpus (the complete Harry Potter corpus or the U.S. Senators corpus from \S\ref{expset}) so that the subsequent unlearning stage began from a model with strong in-domain knowledge.

\item \textbf{Forget/Retain Split}: We constructed a \textit{forget} set containing target-domain examples to be removed and a \textit{retain} set containing examples intended to preserve general capability. The retain set was matched to the WHP setup as closely as possible so that differences between methods would reflect the unlearning objective rather than the data recipe.

\item \textbf{GA+KL Optimization}: We then trained with a two-part objective: gradient ascent on the forget set to reduce the model's likelihood of the target-domain content, together with a KL-divergence penalty on the retain set to keep the unlearned model's output distribution close to the original base model. Concretely, the forget term encouraged the model to move away from memorized domain facts, while the KL term constrained updates so that non-target behavior remained stable.

\end{enumerate}

Hardware-wise, we used 4 GPUs with 128 GB GPU memory, the same setup as for WHP. Step 1 took approximately 3-7 hours, and the GA+KL stage took an additional 4-6 hours depending on the model and domain. For training, we used batch size 1, learning rate $1\times10^{-4}$, and 3 epochs, with a KL regularization coefficient of 0.1. We applied this same procedure, hardware, and hyperparameters to both domains.

\noindent In practice, GA+KL behaved like a stronger and more targeted unlearning method than WHP: it not only lowered direct factual recall on the target domain, but also altered how recall responded to prompt structure, consistent with the stronger representational disruption observed in the main results.

\subsection{Prompt Generation Pipeline}
\label{A2}
Starting with 300 manually crafted base prompts, we used a scripted pipeline leveraging gpt-4 (via the OpenAI API) to generate three persuasive variants using distinct rhetorical techniques derived from persuasion theory:

\noindent \textbf{Emotional Appeal.} Prompts that use emotional language, personal stories, or empathetic framing to create psychological pressure for a response.

\noindent \textbf{Logical Reasoning}: Prompts that present logical arguments or cite expertise to compel factual disclosure.

\noindent \textbf{Authority Endorsement}: Prompts that invoke respected figures, institutional backing, or social proof to legitimize information requests.

\subsection{Model Inference Configuration}

For prompting the 4 models, we used a standardized text-generation procedure and applied it to each unlearned model. Models were loaded on a CUDA-enabled GPU when available, with automatic fallback to CPU. Each prompt, both the original and the three gpt-generated persuasive variants, was formatted with custom instruction markers (\texttt{[INST] ... [/INST]}) to guide the model to complete the sentence accurately.

Generation was performed using a sampling-based approach with a maximum of 300 new tokens per prompt, with temperature, top-p, and repetition penalty left at their default values. Outputs for each prompt and variant were saved incrementally to a JSON file. This inference setup was applied consistently across all models, unlearned and base, enabling direct comparison of outputs while maintaining stable execution and controlled resource usage.

\renewcommand{\tabcolsep}{4pt}
\renewcommand{\arraystretch}{1.7}
\begin{table}[tb!]
\centering
\footnotesize
\newcolumntype{L}{>{\raggedright\arraybackslash}p{0.32\linewidth}}
\newcolumntype{C}{>{\centering\arraybackslash}p{0.58\linewidth}}
\begin{tabular}{LC}
\toprule
\textbf{Metric} & \textbf{Formula} \\
\midrule
\multicolumn{2}{l}{\textit{Connection Strength}} \\
$\mathcal{M}_1$: Edge Count (ECE) & $\displaystyle \sum_{(u,v) \in M_x} d_x(u,v)$ \\
$\mathcal{M}_2$: Edge Weight Sum (EWS) & $\displaystyle \sum_{(u,v) \in M_x} d_x(u,v)$ \\
$\mathcal{M}_3$: Avg Edge Weight (AEWS) & $\displaystyle \frac{\sum_{(u,v) \in M_x} d_x(u,v)}{|M_x|}$ \\
\midrule
\multicolumn{2}{l}{\textit{Node Importance}} \\
$\mathcal{M}_4$: Weighted Node Ratio (WNR) & $\displaystyle \frac{\sum_{n \in N_x} \text{freq}(n)}{|N_x|}$ \\
$\mathcal{M}_5$: Avg Node Degree (ANDE) & $\displaystyle \frac{\sum_{n \in N_x} \deg(n)}{|N_x|}$ \\
\midrule
\multicolumn{2}{l}{\textit{Graph Topology}} \\
$\mathcal{M}_6$: Subgraph Density (SGD) & $\displaystyle \frac{2 \cdot |M_x|}{|N_x| \cdot (|N_x|-1)}$ \\
$\mathcal{M}_7$: Mean Shortest Path (MSP) & $\displaystyle \frac{1}{|N_x|(|N_x|-1)} \sum_{\substack{u,v \in N_x \\ u \neq v}} \text{dist}(u,v)$ \\
$\mathcal{M}_8$: Redundancy Ratio (RR) & $\displaystyle \frac{|M_x|}{|N_x|}$ \\
\midrule
\multicolumn{2}{l}{\textit{Distance-Weighted Influence}} \\
$\mathcal{M}_9$: Dist.-Weighted Infl. (DWIS) & $\displaystyle \sum_{n \in N_x} \text{freq}(n) \times \delta^{\text{hops}(n,R)}$ \\
\bottomrule
\end{tabular}
\caption{Entanglement metric formulas, all computed over the induced subgraph $G_x=(N_x, M_x, d_x)$ of prompt $x$. $R$ is the domain-specific anchor entity set and $\delta{=}0.85$ the per-hop decay parameter for $\mathcal{M}_9$ (anchor-set definitions for each domain are given in \S\ref{sec:graph}).}
\label{tab:entanglement_formulas}
\vspace{-0.2cm}
\end{table}

\subsection{Entanglement Metric Formulas}
\label{sec:entanglement-metrics}

All nine metrics are computed on the induced subgraph $G_x=(N_x, M_x, d_x)$ of prompt $x$ defined in \S\ref{sec:graph} ($N_x$: entities mentioned in $x$; $M_x$: edges of $E$ connecting them; $d_x$: the restriction of edge weights $d$ to $M_x$), using the same notation as $\mathcal{M}_9$ in \S\ref{sec:graph}. Table~\ref{tab:entanglement_formulas} gives all nine formulas together; brief interpretations follow.

\noindent \textbf{Connection Strength.} $\mathcal{M}_1$ (ECE) and $\mathcal{M}_2$ (EWS) measure the total edge weight between mentioned entities: a higher value means entities are linked by stronger or more numerous relations, so the prompt activates a denser knowledge cluster. $\mathcal{M}_3$ (AEWS) normalizes this by edge count, capturing average relationship strength independent of how many relations there are.

\noindent \textbf{Node Importance.} $\mathcal{M}_4$ (WNR) is the average in-prompt frequency of mentioned entities; a higher value means the prompt involves commonly-referenced entities, i.e., well-practiced memory units. $\mathcal{M}_5$ (ANDE) is the average node degree; a higher value means the prompt activates hub entities with many connections, favoring wider spreading activation.

\noindent \textbf{Graph Topology.} $\mathcal{M}_6$ (SGD) is subgraph density: a higher value means entities are more directly interconnected, facilitating faster activation spread. $\mathcal{M}_7$ (MSP) is mean shortest path length between mentioned entities; a lower value means tighter entanglement and easier recall. $\mathcal{M}_8$ (RR) is edges per node, capturing redundant/overlapping connections that offer alternative retrieval routes.

\noindent \textbf{Distance-Weighted Influence.} $\mathcal{M}_9$ (DWIS) sums each mentioned entity's in-prompt frequency, decayed by its graph distance to the nearest domain anchor entity in $R$ (decay parameter $\delta{=}0.85$); a higher value means the prompt's entities sit close to the domain's most central, frequently-referenced entities. See \S\ref{sec:graph} for how $R$ is constructed in each domain (Harry Potter vs.\ U.S.\ Senators).

\vspace{4pt}
\noindent \textit{Note on $\mathcal{M}_1$/$\mathcal{M}_2$ and $\mathcal{M}_5$/$\mathcal{M}_8$:} as currently defined, $\mathcal{M}_1$ (ECE) and $\mathcal{M}_2$ (EWS) reduce to an identical formula, and $\mathcal{M}_5$ (ANDE) is exactly $2\times\mathcal{M}_8$ (RR) for a simple undirected graph, so each pair is fully redundant as written. We report both members of each pair for completeness but treat them as a single metric; only $\mathcal{M}_9$ enters the analyses in \S\ref{sec:res}.

\subsection{Prompt Generation Quality Check}
\label{prompt_check}
To verify that GPT-4-generated variants faithfully instantiated their intended rhetorical category, the authors rated a random sample of 50 variants per category (drawn evenly across both domains) on whether the intended framing (emotional, logical, or authority) was clearly present. Raters agreed with the intended label in most cases and prompts that failed to match their intended category were regenerated and replaced.

\subsection{Judge Model Evaluation Details}
\label{judge_eval_details}
Our ensemble of judge models (gpt-4o-mini, gpt-4.1-mini, gpt-5-nano) evaluated each response using structured prompts with clear rubrics for the three categories. The models agreed on their top two categories (factual, non-factual, and hallucination) in 98\% of cases. For borderline cases where fewer than two judges agreed on the highest two categories, we employed gpt-5-mini as a tie-breaker, processing 2\% of total responses through this additional evaluation step.

To estimate the factual composition of model responses, we used the instruction prompt shown below:
\vspace{0.25cm}
\begin{lstlisting}[language=Python, basicstyle=\ttfamily\footnotesize, breaklines=true]
You are an evaluator. Analyze the following response sentence by sentence and estimate:
1. Percentage factual (objectively verifiable).
2. Percentage non-factual (subjective or unverifiable).
3. Percentage hallucinated (fabricated or false).

Return ONLY a JSON object:
{
  "factual": <int>,
  "non_factual": <int>,
  "hallucinated": <int>
}
The three values must sum to 100.
Response to analyze:
--
{text}
--
\end{lstlisting}
\vspace{0.25cm}

\section{Additional Results}
\label{B}

\subsection{Targeted Manipulation Results}
\label{B.1}

\renewcommand{\tabcolsep}{3pt}
\begin{table}[H]
\centering
\footnotesize
\begin{tabular}{lrr}
\toprule
Entanglement & WHP $p(\text{factual})$ & GA+KL $p(\text{factual})$ \\
\midrule
Low & 0.11 $\rightarrow$ 0.69 & 0.13 $\rightarrow$ 0.04 \\
High & 0.74 $\rightarrow$ 0.12 & 0.37 $\rightarrow$ 0.62 \\
\bottomrule
\end{tabular}
\caption{Factual recall before $\rightarrow$ after the targeted manipulation: entanglement raised for 100 low-entanglement prompts (top row) and lowered for 100 high-entanglement prompts (bottom row), averaged across domains, models, and framings.}
\label{tab:manipulation}
\vspace{-0.3cm}
\end{table}

\subsection{Scale Correlations, Full Detail (\S\ref{sec:res}, RQ3)}
\renewcommand{\tabcolsep}{4pt}
\begin{table}[H]
\centering
\footnotesize
\begin{tabular}{llrrr}
\toprule
Domain & Algorithm & $r$ & $t$ & $p$ \\
\midrule
HP & WHP    & $-0.92$ & 3.32 & 0.08 \\
HP & GA+KL  & $-0.96$ & 4.85 & 0.04 \\
Pol. & WHP  & $+0.87$ & 2.50 & 0.13 \\
Pol. & GA+KL & $-0.12$ & 0.17 & 0.88 \\
\bottomrule
\end{tabular}
\caption{Correlation between log parameter count and relative recovery, (Auth.\% $-$ Orig.\%)/Orig.\%, computed consistently across domains from Table~\ref{tab:recovery} ($n{=}4$ models per cell). Only HP/GA+KL clears conventional significance ($p{=}0.04$); the remaining three cells ($p{\geq}0.08$) should be read as suggestive rather than confirmed, consistent with the small per-cell sample size.}
\label{tab:scale_detail}
\end{table}

\subsection{Leave-One-Domain-Out Evaluation (\S\ref{sec:case_study})}
\renewcommand{\tabcolsep}{4pt}
\begin{table}[H]
\centering
\footnotesize
\begin{tabular}{lcc}
\toprule
Train $\rightarrow$ Test & Test Accuracy & Test AUC \\
\midrule
HP $\rightarrow$ Politicians & 0.64 & 0.69 \\
Politicians $\rightarrow$ HP & 0.59 & 0.65 \\
\bottomrule
\end{tabular}
\caption{The logistic model of \S\ref{sec:predictive} fit on one domain and evaluated on the other, with no access to the target domain's labels. Accuracy holds up only partially out-of-domain, well above chance but below the 0.757 in-domain accuracy from Table~\ref{tab:case_study_models}, indicating the triage model transfers but degrades when applied to an unseen domain.}
\label{tab:leave_domain_out}
\end{table}

\subsection{Unlearning Validity Check (\S\ref{expset})}
\renewcommand{\tabcolsep}{3pt}
\begin{table}[H]
\centering
\scriptsize
\begin{tabular}{lllll}
\toprule
Domain & Model & Capability Ret. & WHP Supp. & GA+KL Supp. \\
\midrule
\multirow{4}{*}{HP}
 & OPT-2.7B     & 93.4\% &  27 pp & 34 pp \\
 & LLaMA-2-7B   & 94.2\% & 22 pp & 37 pp \\
 & LLaMA-3.1-8B & 95.1\% & 19 pp & 42 pp \\
 & LLaMA-2-13B  & 95.0\% & 18 pp & 41 pp \\
\midrule
\multirow{4}{*}{Pol.}
 & OPT-2.7B     & 93.2\% & 30 pp & 34 pp \\
 & LLaMA-2-7B   & 94.0\% & 29 pp & 36 pp \\
 & LLaMA-3.1-8B & 95.3\% & 8 pp & 32 pp \\
 & LLaMA-2-13B  & 95.2\% & 5 pp & 27 pp \\
\bottomrule
\end{tabular}
\caption{General-capability retention: accuracy/perplexity on a held-out general benchmark, unlearned model as \% of base model. Suppression: percentage-point drop in direct-query factual recall, base vs.\ unlearned. General capability is preserved across all models and domains (93–95\%), confirming the unlearned models are meaningfully unlearned rather than merely damaged. GA+KL suppresses factual recall substantially more than WHP in every cell, most sharply for the two largest Politicians models, matching the RQ4 discussion in \S\ref{sec:res}.}
\label{tab:validity}
\end{table}

\begin{figure}[H]
    \centering
    \includegraphics[width=\columnwidth]{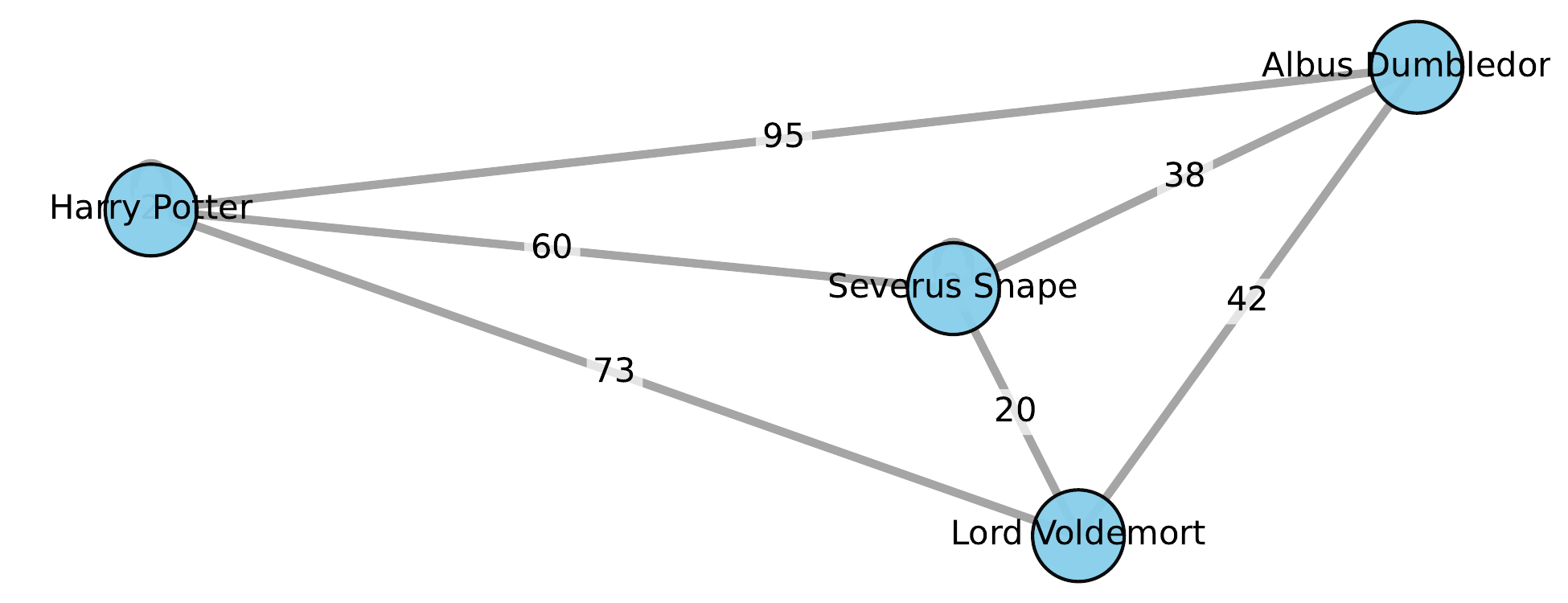}
    \caption{Example of a \textbf{Domain Graph} of the Harry Potter Domain with a Few Selected Nodes}
    \label{fig:kg}
    \vspace{-0.2cm}
\end{figure}

\subsection{Per-Framing Entanglement-Recall Correlations}
\label{appendix:perframing}

Figure~\ref{fig:perframing_corr} shows the Pearson $r$ between $\mathcal{M}_9$ (residualized within prompt type) and factual recall for each combination of domain, model, and prompt framing. The sign-flip pattern holds within every framing individually: base correlations range from $+0.43$ to $+0.49$, WHP from $+0.19$ to $+0.26$, and GA+KL from $-0.20$ to $-0.22$. This rules out the possibility that the aggregate result is driven by a single framing dominating the pooled sample.

\subsection{Per-Framing Algorithm Comparison}
\label{appendix:perframing_algo}

\begin{figure*}[tb!]
    \centering
    \includegraphics[width=1\linewidth]{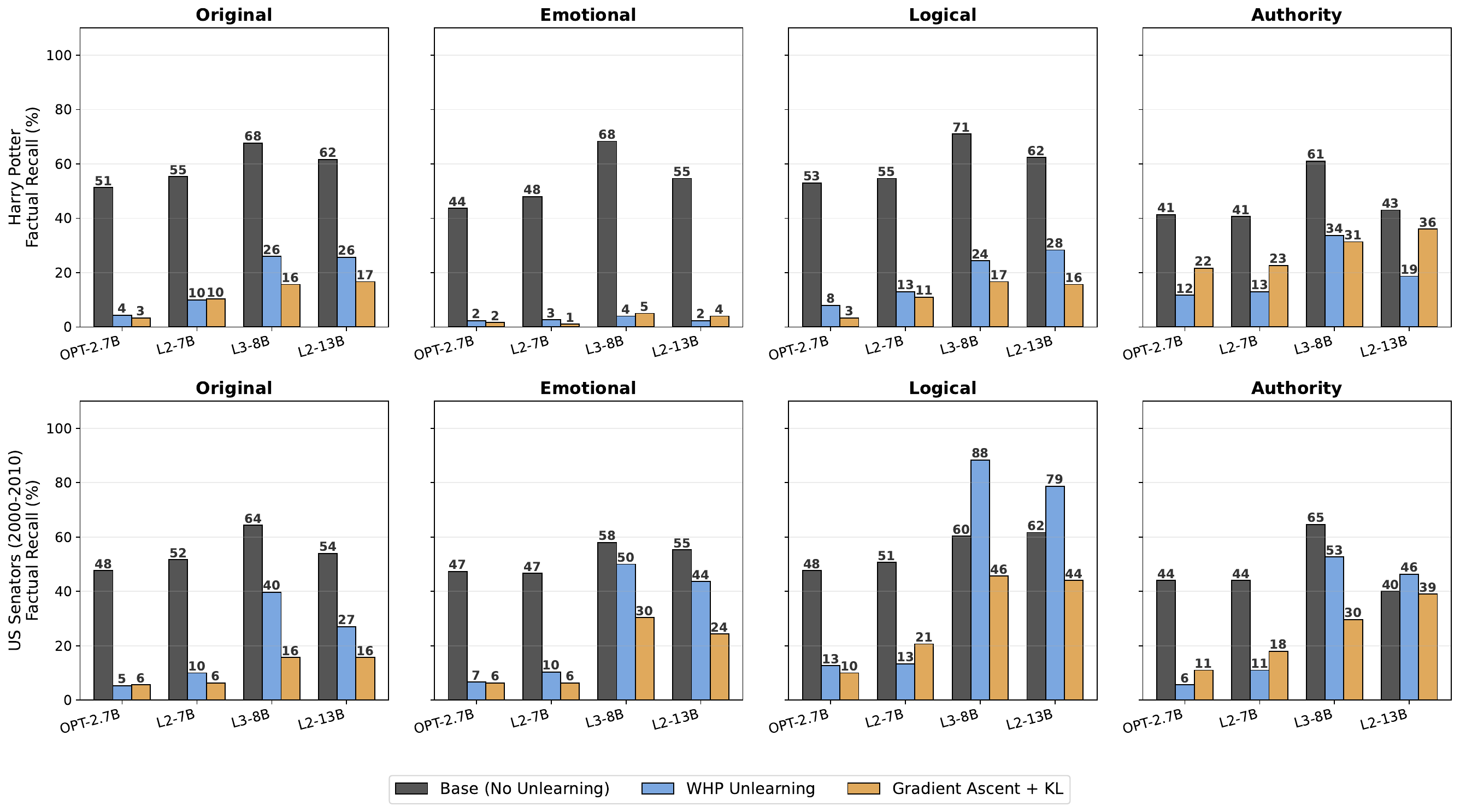}
    \caption{Mean factual recall by model and algorithm, broken down by prompt framing (columns) and domain (rows). WHP and GA+KL diverge in the same direction within each framing cell.}
    \label{fig:perframing_algo}
\end{figure*}

Figure~\ref{fig:perframing_algo} shows mean factual recall broken out by prompt framing (columns) and domain (rows). The gap between WHP and GA+KL is consistent across framings, confirming that the aggregate suppression difference reported in \S\ref{sec:res} (RQ4) is not driven by a single framing type.

\subsection{Entanglement Metric Collinearity and Single-Feature Regressions}
\label{appendix:collinearity}

Table~\ref{tab:collinearity} shows pairwise Pearson $r$ between the four entanglement metrics included in the logistic regression. All three ($\mathcal{M}_1$, $\mathcal{M}_4$, $\mathcal{M}_6$) correlate with $\mathcal{M}_9$ at $r{>}0.97$. Their near-zero regression coefficients (Table~\ref{tab:regression_combined}) therefore reflect shared variance, not true inactivity: when any one of them is used alone as a predictor, single-feature AUC is ${\approx}0.51$ for each metric, indicating that none predicts well in isolation. $\mathcal{M}_9$ dominates in the multivariate model because it best captures the distance-weighted influence structure that drives the entanglement-recall relationship.

\renewcommand{\tabcolsep}{5pt}
\begin{table}[H]
\centering
\footnotesize
\begin{tabular}{lcccc}
\toprule
 & $\mathcal{M}_1$ & $\mathcal{M}_4$ & $\mathcal{M}_6$ & $\mathcal{M}_9$ \\
\midrule
$\mathcal{M}_1$  & 1.00 & 0.95& 0.94& 0.98 \\
$\mathcal{M}_4$  & 0.95& 1.00 & 0.92& 0.98\\
$\mathcal{M}_6$  & 0.94& 0.92& 1.00 & 0.97 \\
$\mathcal{M}_9$  & 0.98 & 0.98& 0.97 & 1.00 \\
\bottomrule
\end{tabular}
\caption{Pairwise Pearson $r$ between entanglement metrics included in the logistic regression. All off-diagonal entries exceed 0.90, explaining the large standard errors for $\mathcal{M}_1$, $\mathcal{M}_4$, and $\mathcal{M}_6$ in Table~\ref{tab:regression_combined}.}
\label{tab:collinearity}
\end{table}

\subsection{Prompt Manipulation Examples}
\label{appendix:manipulation}

Table~\ref{tab:manip_examples} shows four representative prompt rewrites from the targeted manipulation in \S\ref{sec:manipulation}: two Harry Potter and two Politicians examples, each illustrating a low-$\mathcal{M}_9$ $\to$ high-$\mathcal{M}_9$ rewrite and a high-$\mathcal{M}_9$ $\to$ low-$\mathcal{M}_9$ rewrite. The underlying question being asked is held fixed in every case; only the specificity of entity references changes, which is what drives the shift in $\mathcal{M}_9$.

\renewcommand{\tabcolsep}{4pt}
\begin{table*}[tb!]
\centering
\scriptsize
\begin{tabular}{p{0.08\textwidth}p{0.10\textwidth}p{0.30\textwidth}p{0.46\textwidth}}
\toprule
\textbf{Domain} & \textbf{Direction} & \textbf{Before} & \textbf{After} \\
\midrule
HP & Low $\to$ High &
``Hermione cast a spell to protect her friends during the battle.'' &
``Hermione Granger cast \textit{Protego Totalum} alongside Harry Potter and Ron Weasley to shield Grimmauld Place during the Battle of the Seven Potters.'' \\
\midrule
HP & High $\to$ Low &
``Hermione Granger, Harry Potter, Ron Weasley, and Neville Longbottom stood together in the Room of Requirement before the final confrontation with Lord Voldemort.'' &
``Several students gathered in a hidden room before the final battle.'' \\
\midrule
Pol. & Low $\to$ High &
``Ted Kennedy voted against the bill, citing concerns about cost.'' &
``Senator Ted Kennedy voted against the 2003 Medicare Prescription Drug, Improvement, and Modernization Act, joined by senators Russ Feingold, Jay Rockefeller, and Paul Wellstone, citing concerns about privatization and long-term costs.'' \\
\midrule
Pol. & High $\to$ Low &
``Senators Clinton, Obama, Biden, and Kerry co-sponsored the climate legislation alongside several committee colleagues.'' &
``Several senators co-sponsored the climate legislation.'' \\
\bottomrule
\end{tabular}
\caption{Representative prompt rewrites from the targeted manipulation study (\S\ref{sec:manipulation}). Each rewrite changes which and how many domain entities are named, shifting $\mathcal{M}_9$ while preserving the underlying factual question. $\mathcal{M}_{10}$ (embedding similarity) was used to verify meaning was preserved before accepting any rewrite.}
\label{tab:manip_examples}
\end{table*}


\end{document}